\documentclass{article}

\usepackage{microtype}
\usepackage{graphicx}
\usepackage{subfigure}
\usepackage{booktabs} % for professional tables

\usepackage{hyperref}

\usepackage[accepted]{icml2024}

\usepackage{amsmath}
\usepackage{amssymb}
\usepackage{mathtools}
\usepackage{amsthm}

\usepackage[capitalize,noabbrev]{cleveref}

\theoremstyle{plain}

\theoremstyle{definition}

\theoremstyle{remark}

\usepackage{amsmath,amsfonts}
\usepackage{algorithmic}
\usepackage{graphicx}
\usepackage{textcomp}
\usepackage{xspace}
\usepackage{xcolor}
\usepackage{subfigure}
\usepackage{threeparttable}
\usepackage{booktabs}
\usepackage{url}
\usepackage{multirow} 
\usepackage{pifont}
\usepackage{algorithm}
\usepackage{algorithmic}
\usepackage{makecell}
\usepackage{newfloat}
\usepackage{listings}
\usepackage{etoolbox}
\usepackage{latexsym}
\usepackage{times}
\usepackage{xcolor}
\usepackage{colortbl}
\usepackage{newfloat}
\usepackage{listings}
\usepackage{newfloat}
\usepackage{listings}

\usepackage{amssymb}
\usepackage{pifont}
\usepackage{graphicx}
\usepackage{latexsym}
\usepackage{times}
\usepackage{soul}
\usepackage{url}
\usepackage{booktabs}
\usepackage{algorithm}
\usepackage{algorithmic}
\usepackage{multirow}
\usepackage{enumitem}

\def\model{MH-pFLID}

\usepackage[textsize=tiny]{todonotes}

\icmltitlerunning{Submission and Formatting Instructions for ICML 2024}

\begin{document}

\twocolumn[
\icmltitle{MH-pFLID: Model Heterogeneous personalized Federated Learning via Injection and Distillation for Medical Data Analysis}



\icmlsetsymbol{equal}{\dag}
\renewcommand{\thefootnote}{\fnsymbol{footnote}}
\begin{icmlauthorlist}
\icmlauthor{Luyuan Xie}{yyy,xxx,zzz}
\icmlauthor{Manqing Lin}{yyy,xxx,zzz}
\icmlauthor{Tianyu Luan}{comp,equal}%\footnote{}
\icmlauthor{Cong Li}{yyy,xxx,zzz}
\icmlauthor{Yuejian Fang}{yyy,xxx,zzz}
\icmlauthor{Qingni Shen}{yyy,xxx,zzz}
\icmlauthor{Zhonghai Wu}{yyy,xxx,zzz}
\end{icmlauthorlist}

\icmlaffiliation{yyy}{School of Software and Microelectronics, Peking University, Beijing, China.}

\icmlaffiliation{xxx}{National Engineering Research Center for Software Engineering, Peking University, Beijing, China.}

\icmlaffiliation{zzz}{PKU-OCTA Laboratory for Blockchain and Privacy Computing, Peking University, Beijing, China.}
\icmlaffiliation{comp}{State University of New York at Buffalo, NY, United States.}

%Corresponding author
\icmlcorrespondingauthor{Tianyu Luan}{tianyulu@buffalo.edu}

\icmlkeywords{Machine Learning, ICML}

\vskip 0.3in
]



\printAffiliationsAndNotice{} % otherwise use the standard text.

\begin{abstract}
Federated learning is widely used in medical applications for training global models without needing local data access. However, varying computational capabilities and network architectures (system heterogeneity), across clients pose significant challenges in effectively aggregating information from non-independently and identically distributed (non-IID) data. 
Current federated learning methods using knowledge distillation require public datasets, raising privacy and data collection issues. Additionally, these datasets require additional local computing and storage resources, which is a burden for medical institutions with limited hardware conditions. In this paper, we introduce a novel federated learning paradigm, named Model Heterogeneous personalized Federated Learning via Injection and Distillation (\model{}). Our framework leverages a lightweight messenger model that carries concentrated information to collect the information from each client. We also develop a set of receiver and transmitter modules to receive and send information from the messenger model, so that the information could be injected and distilled with efficiency.
% for each client to separate local biases from generalizable information, reducing biased data collection and mitigating client drift. 
Our framework eliminates the need for public datasets and efficiently share information among clients.
Our experiments on various medical tasks including image classification, image segmentation, and time-series classification, show \model{} outperforms state-of-the-art methods in all these areas and has good generalizability.
\end{abstract}

\begin{figure}[t]
%\vskip 0.2in
\includegraphics[width=1\linewidth]{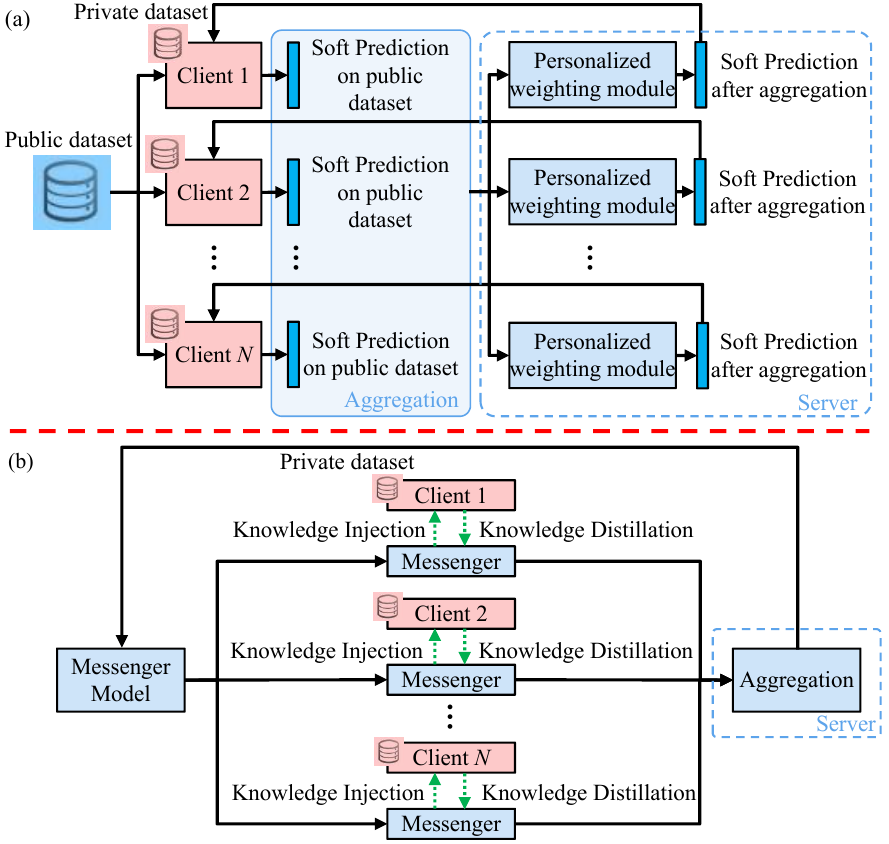}
\centering
\caption{(a) Previous methods such as KT-pFL~\cite{NEURIPS2021_5383c731} require an extensive public dataset to generate soft predictions to carry information from local clients to the server. These methods are highly limited by the high cost and privacy concerns of medical public datasets in real applications. (b) Our new framework \model{} does not require such a public dataset for training. We use a lightweight messenger model to carry and transform the information among model heterogeneous clients and the server.}
\label{fig_1}
%\vskip -0.2in
\end{figure}

\section{Introduction}
\label{submission}
% Vanilla 
% Federated learning finds widespread application in medical scenarios due to its ability to collaboratively train a global model without direct access to the local data of each healthcare institution.
% While federated learning demonstrates its capability to protect the privacy of patient data, it still grapples with the challenges of system heterogeneity and statistic heterogeneity. Specifically, the potential adoption of customized model structures by each institution and the non-Independently and Identically Distributed (Non-IID) nature of local data across different healthcare institutions pose significant challenges. The trained global model may not effectively deal with these issues. The non-independently and identically distributed (non-IID) \cite{9743558} nature of local data across different healthcare institutions offers the potential for better performance and generalizability in the overall model, but it also increases the complexity of information aggregating.

Federated learning \cite{fedavg} is extensively used in medical applications for its capability to train a global model collaboratively without requiring direct access to local data from each healthcare institution. Considering the local data of different healthcare institutes is mostly non-independent and identically distributed (non-IID), previous works such as \cite{fedrep,apfl,lg-fedavg} have focused on solving the statistic heterogeneity problems, in which the data distribution is the major difference among clients.
However, system heterogeneity, such as the model architecture differences among clients, is also a commonly existing situation in real-world applications. The diverse requirements, hardware resource concerns, and regulatory risks lead different hospitals to choose and maintain different model architectures, and different model architectures result in different fitting abilities, different performance behaviors, and difficulties in model aggregation. The broad usage and problem complexity make the model heterogeneity problem a crucial and challenging task.

As illustrated in \cref{fig_1}(a), the existing methods addressing statistic and system heterogeneity are based on Knowledge Distillation \cite{hinton2015distilling}. These approaches involve the exchange of soft predictions on a public dataset among different clients, enabling the transfer of knowledge from one client to another (e.g., FedMD~\cite{DBLP:journals/corr/abs-1910-03581}, FedDF~\cite{NEURIPS2020_18df51b9}, KT-pFL~\cite{NEURIPS2021_5383c731}). Although these methods have made progress in addressing system and statistic heterogeneity, they would still rely on a prerequisite public dataset to generate soft predictions. The high privacy concerns and the intricate process of collecting and publishing medical data restrict the practical applicability of these methods. Moreover, the large scale of public datasets makes local training challenging. It is uneasy for many healthcare institutions to acquire the computational resources for training on large public datasets. This could be even more burdensome considering these extra required computational resources are not needed during inference time \cite{baltabay2023designing}.

% To address these problems, we propose a Federated Learning framework based on the Teacher-Student and parameter decouplin paradigm, as illustrated in Figure 1 (b). This framework enables the training of heterogeneous models for each client without the need for server-side computations or a public dataset. Within each participating institution in Federated Learning, we insert the same lightweight model, where the lightweight model's parameters and FLOPs are significantly smaller than the local heterogeneous models. We designate the local Model Heterogeneousas the Teacher model and the inserted lightweight model as the Student model. Through knowledge distillation, the local Model Heterogeneoustransfers knowledge to the lightweight model. Subsequently, we aggregate the uploaded Student models to obtain a global knowledgeable Student model, which guides the training of the Teacher model during local training to enhance its performance.

%previous round of aggregated

To eliminate the reliance on public datasets, we propose a novel injection-distillation paradigm to address the challenges of heterogeneous models under the distribution of non-IID data. Unlike traditional approaches that rely on soft prediction generated from public datasets, our method utilizes an extremely lightweight \textit{messenger} model for information transfer. Our paradigm consists of three steps: knowledge injection, knowledge distillation, and aggregation. We insert the messenger model into each local client. During the knowledge injection phase, the knowledge from the messenger is injected into each local model. During the knowledge distillation phase, the client’s knowledge is distilled into the messenger model while training on local data. In the knowledge aggregation phase, we aggregate the knowledge by combining the parameters of the messenger models. Additionally, the small parameter size of our messenger model ensures that its local training imposes minimal additional burden compared to local training on the local dataset. This additional load is less than what would be required for local training on a public dataset.

We name our framework as \textit{Model Heterogeneous personalized Federated Learning via Injection and Distillation (\model{})}. To facilitate efficient information transfer among heterogeneous models using a compact messenger model, we conceptualize the messenger model as a set of information bases. This approach is more efficient than directly embedding information in the messenger, while still allowing for the recovery of information acquired from local data. The messenger model comprises a codebook for storing globally shared information and a head for supervised training. 
In the injection phase, an attention-based, client-specific information \textit{receiver} is employed for each client to receive information from the messenger. This receiver gets client-specific biases from global information, enabling the model to get other client information within the receiver and receive the generalizable information in the codebook. Similarly, during the distillation phase, a client-specific information \textit{transmitter} is employed to relay information from the local model to the codebook. This disentanglement design significantly enhances the representation capacity of the messenger model and reduces biased information collected from each client, thus mitigating the issue of client drift.

% {\color{green}In the distillation phase, an attention-based, client-specific \textit{information transmitter} is employed to relay information from the local model to the codebook. This transmitter separates client-specific biases from global information, enabling the model to retain local biases within the transmitter and transmit the generalizable information in the codebook. Similarly, during the injection phase, a client-specific "information receiver" is employed for each client to receive information from the messenger. This disentanglement design significantly enhances the representation capacity of the messenger model and reduces biased information collected from each client, thus mitigating the issue of client drift.}

In summary, our contributions include the following:
\begin{itemize}[noitemsep,topsep=0pt]
% \item To the best of our knowledge, we are the first to propose a Federated Learning approach based on the Teacher-Student paradigm and parameter decoupling for personalized heterogeneous models. Compared to other methods, \model{} effectively utilizes parameter aggregation to acquire knowledge from other clients, thereby improving performance. 
\item We propose a  personalized federated learning method for heterogeneous models based on an injection-distillation paradigm named \model{}. Compared to previous methods, \model{} eliminates reliance on public medical datasets and the associated extra costs of local training.
\item We design a lightweight \textit{messenger} model to transfer information among different clients. The messenger can efficiently integrate information from each client via a codebook-style design, while barely increasing local training costs.
\item We design a set of \textit{transmitter} and \textit{receiver} modules for each client to disentangle local biases from generalizable information. This disentanglement effectively reduces the biased information gathered from each client, thereby addressing and reducing the problem of client drift.
\item We validate our \model{} on multiple medical tasks, including medical image classification, medical image segmentation, and medical time series classification. The experimental results show that our method outperforms state-of-the-art results on all these tasks, proving its effectiveness and generalizability in various medical applications.

% \item We validated that \model{} is applicable to multiple medical tasks, including three medical image classification tasks, one medical image segmentation task, and one medical signal classification task. The experimental results show that our method outperforms state-of-the-art results, proving its generality in various medical tasks within the federated paradigm.
\end{itemize}

\begin{figure*}[ht]
\vskip 0.2in
\includegraphics[width=0.82\linewidth]{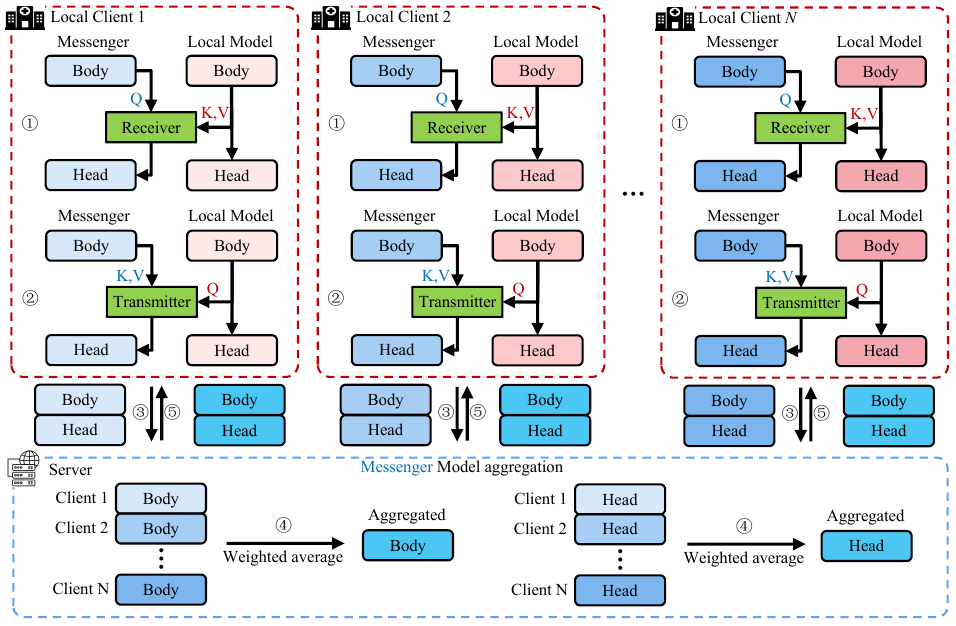}
\centering
\caption{Overview of our proposed \model{} framework. Each training cycle consists of 5 steps. From 1 to 5: \textcircled{1} Knowledge injection stage. We design an Information Receiver module to utilize the aggregated information in Messenger to train the local model. \textcircled{2} Knowledge distillation stage. We design an Information Transmitter module to transmit the personalized information from the local model to the messenger. \textcircled{3} Uploading the messenger parameters on each client to the server. \textcircled{4} Messenger aggregation on the server using a weighted average strategy. \textcircled{5} Downloading the aggregated messenger parameters to each server. More details can be found in \cref{sec:pipeline} and \cref{sec:irit}.}
\label{fig_2}
\vskip -0.2in
\end{figure*}
%FedAvg performance will significantly decrease. Some existing methods as  have proposed various regularization methods to reduce the impact of statistic heterogeneity. However,
\section{Related Works}
\subsection{Personalized Federated Learning in Statistic Heterogeneity}
For classic federated learning algorithms such as the FedAvg \cite{fedavg}, SCAFFOLD \cite{karimireddy2020scaffold} and FedProx \cite{fedprox}, aiming to train a single global model across all clients. Some work has shown that when faced with statistic heterogeneity issues, it is difficult for a single model to have good performance in all clients. Personalized Federated learning is proposed to train a personalized model for each client to effectively alleviate the above problems. It includes methods such as clustering \cite{clustered1,clustered2}, model interpolation\cite{li2021ditto,apfl,lcfed}, multi-task learning \cite{li2021ditto,mtl1,mtl2,mtl3}, local memorization \cite{pmlr-v162-marfoq22a} and parameter decoupling \cite{fedrep,lg-fedavg,fedper,fedhealth,knnper}.

 %APFL \cite{apfl} adopts the joint prediction mode which is a mixture of global model and local model with adaptive weight. LG-FedAvg \cite{lg-fedavg} and FedRep \cite{fedrep} use parameter decoupling to jointly learn the global part and local part of the client model, while only the global part was sent to the server. The difference between them is the definition of which part of the model is the global part. kNN-Per \cite{knnper} uses the output of the K global representations closest to the global representation of the input to be evaluated to make predictions. Ditto \cite{li2021ditto} adds a regular term to the original local loss function of each client to measure the deviation between the local model and the global model.

\subsection{Personalized Federated Learning in System Heterogeneity}
%The main reason for system heterogeneity is the difference in storage and computing capabilities among different clients or institutions. This will lead to different clients or institutions customizing specific structural models based on hardware. Vanilla federated learning or classical personalized federated learning both require consistency in the parameters of each client model, making these methods unsuitable for application in heterogeneous system scenarios.   FedMD \cite{DBLP:journals/corr/abs-1910-03581} exchanges soft predictions between the clients and the server for training personalized heterogeneous models.  DS-pFL \cite{9392310} uses a semi-supervised distillation method to complete training on unlabeled public datasets.

In recent years, some studies used Knowledge Distillation (KD) \cite{hinton2015distilling} to solve the problem of system heterogeneity. The principle is to aggregate local soft predictions in the server such as FedMD \cite{DBLP:journals/corr/abs-1910-03581} and DS-pFL \cite{9392310}. To further address the issues of statistic heterogeneity and system heterogeneity, FedDF \cite{NEURIPS2020_18df51b9} adopts ensemble distillation methods to train heterogeneous models in the server. KT-pFL \cite{NEURIPS2021_5383c731} trains personalized soft prediction weights on the server side to further improve the performance of heterogeneous models. The above methods have made good progress in the study of system heterogeneity. However, their commonality is the need for public datasets and central computing burden. These commonalities limit the use of these methods in medical scenarios.
pFedES \cite{DBLP:journals/corr/abs-2311-06879} and FedLoRA \cite{DBLP:journals/corr/abs-2310-13283} introduced additional parameters to guide local model training, alleviating some commonalities.
%\model{} does not require public datasets and model training in server. These allow \model{} to have a wider range of applications.%\model{} utilizes knowledge injection and distillation to achieve Model Heterogeneoustraining. 

\section{Method}
\subsection{Problem Formulation}
We aim to train a personalized model $f_i(\theta _i; x) $ for client $i$, where $\theta _i$ is the parameters of $f_i$ and $x$ is model input. Each model $f_i$ can only access its own private dataset $D_i = \{x_{ij}, y_{ij}\}$, where $x_{ij}$ is the $j$th input data in $D_i$, and $y_{ij}$ is the $j$th label. We train all $f_i$ collaboratively so that each model can use the information from other clients' datasets without directly accessing data. In real medical scenarios, the data distribution of each client is usually non-IID (statistic heterogeneity). \model{}’s paradigm can be expressed as: % \TL{Should $f(\cdot)$ be personalized？}
%\begin{small} 
\begin{equation} 
% g = G(f({\theta _1};{D_1}), \cdot  \cdot f({\theta _j};{D_i}) \cdot  \cdot ,f({\theta _M};{D_M}))\mathop {}\nolimits^{} 1 < j \le M
\mathbb{G} = \bigcup^N_i{f_i(\theta _i; x)},
\end{equation} 
%\end{small}
where $\mathbb{G}$ represents the the set of $f_i$. N is the total number of participating clients. Due to each client adopting a customized model architecture, the model structure $f_i$ of each client is different (system heterogeneity). So \model{} simultaneously faces two major challenges: statistic heterogeneity and system heterogeneity across $N$ different clients.

\subsection{Pipeline}
\label{sec:pipeline}

The pipeline of \model{} is shown in \cref{fig_2}. Both our local and messenger models are divided into a body model to extract features, and a head model to generate the network output using the features. Our training process consists of 5 steps: \textcircled{1} Knowledge injection, \textcircled{2} Knowledge distillation, \textcircled{3} Uploading messenger models to the server, \textcircled{4} Messenger aggregation, and \textcircled{5} download messenger information back to the server. In the rest of the section, we will explain each step in detail.

% The training stage of \model{} is shown in \cref{fig_2}. In local medical institutions, the model architecture consists of two parts: the messenger model and the local model. The local model represents the customized heterogeneous models adopted by local institutions, which are retained locally in the training stage without parameter sharing. The messenger model is the lightweight model we inserted. It includes messenger body (MB) and head (MH). Considering the computational and local storage issues, the number of messenger model parameters and GFLOPS we inserted is much smaller than that of the local model. The messenger model structure in each institution is consistent. 

%previous round of aggregate
%we freeze the messenger model and use the aggregated messenger model from the previous round to guide the local model in training.

% During the training phase, the messenger model injects the global knowledge into the local model. Then, the trained local model transfers local knowledge to messenger models through distillation methods. Finally, messenger models are uploaded and aggregated, and the aggregated messenger model is distributed to the next training round of communication. A complete communication phase of \model{} is shown in \cref{fig_2}, including 5 processes: \textcircled{1}\textcircled{2}\textcircled{3}\textcircled{4}\textcircled{5}. In local training, there are two stages: \textcircled{1} and \textcircled{2}. We call them Knowledge Injection and Knowledge Distillation, respectively. In the Knowledge Injection stage,
\textbf{Knowledge injection.} The knowledge injection stage is designed to inject information from the messenger to the local model. Specifically, we freeze the messenger model and use the messenger model to guide the local model in training. For client $i$, its knowledge injection stage training loss function $\mathcal{L}_{\textit{inj},i}$ is:
\begin{equation} 
{\mathcal{L}_{\textit{inj},i}} = \lambda^l_{\textit{inj}}\sum\limits_{j = 1}^M {\mathcal{L}^l_{\textit{inj}}(\hat{y}^l_{ij},y_{ij})}  + {\lambda^m_{\textit{inj}}}\sum\limits_{j = 1}^M {\mathcal{L}^m_{\textit{inj}}(\hat{y}^m_{ij},y_{ij})}.
% {\mathcal{L}_{i,stage1}} = {\lambda _1}\sum\limits_{j = 1}^M {\mathcal{L}_{\textit{inj,loc}}}  + {\lambda _2}\sum\limits_{j = 1}^M {\mathcal{L}(\hat{y}_{m,j}^{(i)},y_j^{(i)})} 
\label{eq:l_inji}
\end{equation} 
% {
% \color{blue}
% \begin{equation} 
% \hat{y}^m_{ij} = M_h(R(M_b(x_{ij}), L_b(x_{ij})))
% \end{equation} 
% }
$M$ represents the total number of local data. $\hat{y}^l_{ij}$ and $\hat{y}^m_{ij}$ are the predictions of the local model and messenger model for the $j$-th data of the local client. 
$\mathcal{L}^l_{\textit{inj}}$ and $\mathcal{L}^m_{\textit{inj}}$  represent the loss functions of the local model and the messenger model, respectively. $\lambda^l_{\textit{inj}}$ and $\lambda^m_{\textit{inj}}$ are their corresponding weights. $y_{ij}$ is the label of $x_{ij}$. Here, the local model output $\hat{y}^l_{ij}$ can be defined as
\begin{equation} 
\hat{y}^l_{ij} = L_h(L_b(x_{ij})),
\label{eq:yl1}
\end{equation}
where $L_b(\cdot)$ and $L_h(\cdot)$ are the body and head of the local model, respectively. The messenger model output $\hat{y}^m_{ij}$ can be represented as
\begin{equation} 
\hat{y}^m_{ij} = M_h(R(M_b(x_{ij}), L_b(x_{ij}))),
\label{eq:ym1}
\end{equation}
where $M_b(\cdot)$ is the messenger body network in Fig.2, $L_b(\cdot)$ is the local model body, $R(\cdot)$ is our designed receiver module to receive information from the messenger model (See \cref{sec:irit} for details), and $M_h(\cdot)$ is the messenger head. During the knowledge injection stage, since the messenger model is fixed, the loss can only generate gradients on $L_b(\cdot)$ and $L_h(\cdot)$ via the first term of \cref{eq:l_inji}, and on $L_b(\cdot)$ and $R(\cdot)$ via the second term of \cref{eq:l_inji}.

%} + {\lambda _2} = 1$. $L$ represents the loss function under this federated learning task, such as using cross-entropy ($CE$) Loss in classification tasks. $y_{j,lp}^{(i)}$ and $y_{j,gp}^{(i)}$ represent the prediction results of the local model and the global model, respectively. Considering that local models can better learn global knowledge, we propose an Information Receiver (IR) to better match local features with global features. (See Section 3.3 for details).

\textbf{Knowledge distillation.} The knowledge distillation stage is designed to distill information from the local model to the messenger. Specifically, we freeze the local model and perform knowledge distillation on the messenger model, where the loss function $\mathcal{L}_{\textit{dis},i}$ is represented as:
\begin{equation} 
{\mathcal{L}_{\textit{dis},i}} = \lambda^m_{\textit{dis}}\sum\limits_{j = 1}^M \mathcal{L}^m_{\textit{dis}}(\hat{y}^m_{ij},y_{ij})  + {\lambda^{con}_{\textit{dis}}}\sum\limits_{j = 1}^M {\mathcal{L}^{con}_{\textit{dis}}(\hat{y}^m_{ij},\hat{y}^l_{ij})}.
% {\mathcal{L}_{i,stage1}} = {\lambda _1}\sum\limits_{j = 1}^M {\mathcal{L}_{\textit{inj,loc}}}  + {\lambda _2}\sum\limits_{j = 1}^M {\mathcal{L}(\hat{y}_{m,j}^{(i)},y_j^{(i)})} 
\label{eq:l_diff}
\end{equation} 
$\mathcal{L}^m_{\textit{dis}}$ and $\mathcal{L}^{con}_{\textit{dis}}$ represent the loss function for training the messenger model and the knowledge distillation loss function, respectively. 
For knowledge distillation loss, we use KL divergence to constrain the output of the messenger head and local head model to be under the same distribution, so that the knowledge can be distilled from the local model to the messenger model.
$\lambda^l_{\textit{dis}}$ and $\lambda^{con}_{\textit{dis}}$ are their corresponding weights. Other variables are defined the same as in \cref{eq:l_inji}. Here, the local model output $\hat{y}^l_{ij}$ is defined the same as \cref{eq:yl1}. The messenger model output $\hat{y}^m_{ij}$ can be represented as
\begin{equation} 
\hat{y}^m_{ij} = M_h(T(L_b(x_{ij}),M_b(x_{ij}))),
\end{equation}
where $M_b(\cdot)$, $L_b(\cdot)$, and $M_h(\cdot)$ are defined the same as in \cref{eq:ym1}. $T(\cdot)$ is our designed transmitter module to send information from the local model to the messenger (See \cref{sec:irit} for details). 
During the training of knowledge distillation stage, we try to use both ground-truth $y_{ij}$ and local model output
$\hat{y}^l_{ij}$ to supervised $\hat{y}^m_{ij}$ together, so that the knowledge can be distilled to messenger model. 
Since the local model is fixed, the loss generates gradients on $M_b(\cdot)$ $M_h(\cdot)$, and $T(\cdot)$ via the first and second term of \cref{eq:l_diff}.

%Wherein, ${\lambda _3} + {\lambda _4} = 1$. $L{_d}$ represents the distillation loss function, such as Mean Squared Error (MSE), etc. Considering that global part can better learn local knowledge, we propose an Information Transmitter (IT) to better match global features with local features. the current round of

\textbf{Messenger upload, aggregation, and download.} After training, the messenger is uploaded to the server. Then aggregation of model parameters is performed separately for the body and head of the messenger. The aggregation operation used is the weight averaging from ~\cite{fedavg}, which involves adding up all parameters and dividing them together. Finally, the aggregated model is downloaded and distributed for the next round of training. 

During the inference phase, we directly utilize the well-trained local heterogeneous models for inference. Compared to other existing methods, \model{} eliminates the need for a public dataset. It only inserts lightweight messenger models locally. This enhances the application of model heterogeneous federated learning in medical scenarios.
%, with parameters and computational load much smaller than the local heterogeneous models
% (1) We eliminate the need for public dataset: \model{} does not require a public dataset. It only inserts lightweight messenger models locally. This enhances the application of model heterogeneous federated learning in medical scenarios.
% (2) No reduce computational burden for server: \model{} does not require the server to train personalized prediction weights. The server is involved in model aggregation, reducing the computational load on server.

% (3) Considering the better implementation of Knowledge Injection and Knowledge Distillation stages, we propose information transmitter and information receiver for knowledge transfer of local and messenger features.

%(4) Model Knowledge Distillation for Diverse Tasks: Unlike other methods that use soft predictions for knowledge distillation, \model{} employs knowledge distillation directly on the model. This extends the applicability of \model{} beyond classification tasks to various tasks such as segmentation. This underscores \model{} as a versatile federated learning framework.

\begin{figure}[ht]
\includegraphics[width=0.8\linewidth]{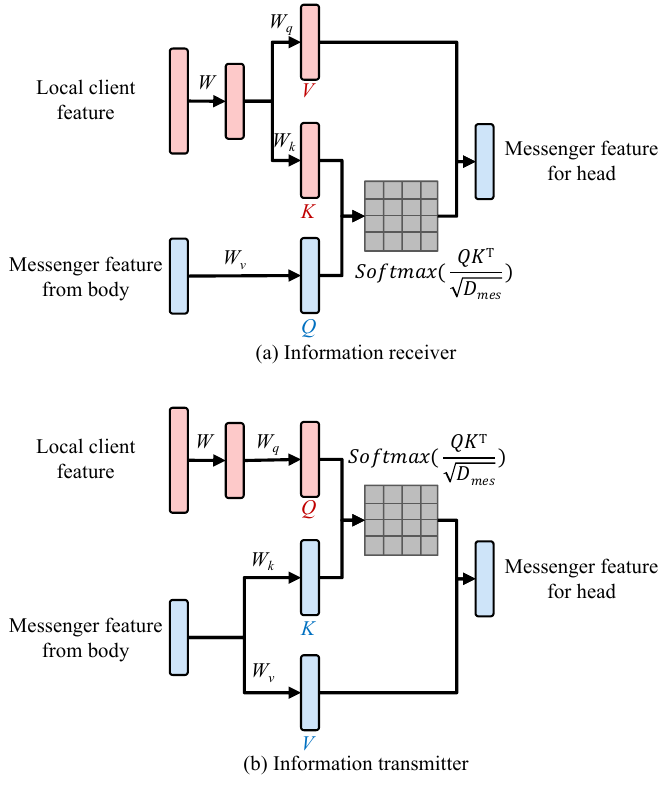}
\centering
\caption{The structure of information receiver (a) and information transmitter (b).}
\label{fig_3}
\end{figure}

\subsection{Information Receiver and Information Transmitter}%%%chongxie
\label{sec:irit}

%Models with different architecture designs focus on disparate feature regions, leading to inconsistent areas of attention for the distilled student models. Solely using the aggregated student models to guide the training of heterogeneous models may result in incorrect guidance for local heterogeneous models due to client drift issues. Based on this, we have designed a feature adapter to personalize the enhancement of features in the global part, thereby addressing the drift issues introduced by the aggregated student models.
Information receiver and transmitter are designed to effectively communicate between local models and the messenger. To achieve a lightweight messenger model, our receiver and transmitter are designed so that the messenger would only need to carry a codebook of the information distilled from local models. The receiver module combines the information from the messenger codebook and injects it into the local model, while the transmitter module decomposes the information from local model and distills it into the messager.
% The information receiver and information transmitter are both attention-based model. The information receiver uses the features extracted from the local body as a codebook, while the features extracted from the messenger body provide a combination of codebook. The information transmitter uses the features extracted from the messenger body as a codebook, while the features extracted from the local body provide a combination of codebook. The messenger model only needs to transport information bases, not full information. This improves the efficiency of transportation and makes the design of messenger model lighter.

\textbf{Information receiver.} The information receiver can be defined as $I_{\textit{loc},R} = R(I_{\textit{loc}}, I_{\textit{mes}})$, where $I_{\textit{loc}}$ and $I_{\textit{mes}}$ are the output feature of local and messenger body, respectively. $I_{\textit{loc},R}$ is the output of the receiver, which is a weighted combination of local body features.
In the knowledge injection stage, we design the information receiver to better match local features with global features. It can enable local models to better receive global knowledge. The information receiver, as illustrated in \cref{fig_3}(a), involves the initial generation of local client features, denoted as $I_{\textit{loc}}$, which undergoes upsampling or downsampling via a linear layer $W_d$ to create features $I^{\prime}_{\textit{loc}}$ with the same dimensions $D_{\textit{mes}}$ as ${I_\textit{mes}}$.
\begin{equation} 
I^{\prime}_{\textit{loc}} = W_d({I_{\textit{loc}}}).
\end{equation} 
% \end{small}
${I_\textit{mes}}$ and ${I^{\prime}_{\textit{loc}}}$ are used to generate Query feature $Q$, Key feature $K$ and Value feature $V$ through $W_k$, $W_q$ and $W_v$. The Query feature $K$ and Key feature $Q$ undergo matrix multiplication to generate the confusion matrix $M_{R}$ as:
\begin{equation}
\begin{array}{c}
     Q = {W_q}({I_\textit{mes}}),
     K = {W_k}(I^{\prime}_{\textit{loc}}),
     V = {W_v}(I^{\prime}_{\textit{loc}}). \\
     M_{R} = \textit{Softmax}(\frac{{Q{K^{\top}}}}{{\sqrt {{D_{\textit{mes}}}} }}).
\end{array}
\end{equation} 
% \begin{small} 
% \begin{equation} 
% M_{R} = \textit{Softmax}(\frac{{Q{K^T}}}{{\sqrt {{D_{\textit{mes}}}} }})
% \end{equation} 
% \end{small}
Finally, the confusion matrix $M_{R}$ is used to perform matrix multiplication with $V$, generating local features ${I_{loc,R}}$ after Information Receiver:
%\begin{small} 
\begin{equation} 
{I_{\textit{loc},R}} = M_{R}{V}.
\end{equation} 
%\end{small}

\textbf{Information transmitter.} The information transmitter can be defined as $I_{\textit{mes},T} = T(I_{\textit{mes}}, I_{\textit{loc}})$, where $I_{\textit{mes},T}$ is the output of the transmitter, which is a weighted combination of messenger body features, shown in \cref{fig_3}(b).
Similar to the knowledge injection stage, in the knowledge distillation stage, we allow global features ${I_\textit{mes}}$ to learn the knowledge of the processed local features $I^{\prime}_{\textit{loc}}$. ${I_\textit{mes}}$ and $I^{\prime}_{\textit{loc}}$ are used to generate Query feature $Q$, Key feature $K$ and Value feature $V$ through $W_k$, $W_q$ and $W_v$. The Query feature $Q$ and Key feature $K$ undergo matrix multiplication to generate the confusion matrix $M_{T}$.
\begin{equation}
\begin{array}{c}
     Q = {W_q}(I^{\prime}_{\textit{loc}}),
     K = {W_k}({I_\textit{mes}}),
     V = {W_v}({I_\textit{mes}}).\\
     M_{T} = Softmax(\frac{{Q{K^{\top}}}}{{\sqrt {{D_\textit{mes}}} }}).
\end{array}
\end{equation} 
Finally, the confusion matrix $M_{T}$ is used to perform matrix multiplication with ${I_\textit{mes}}$, generating global features ${I_{\textit{mes},T}}$ after Information
Transmitter:
\begin{small} 
\begin{equation} 
{I_{\textit{mes},T}} = M_{T}{V}.
\end{equation} 
\end{small}

At inference time, we only need the local model. The messenger model, transmitter, and receiver will not participate in inference.

\begin{table*}[ht]
  \centering
  \caption{The results of classification task in different resolutions. The x2↓, x4↓, and x8↓ are downsampling half, quarter, and eighth of high-resolution images. We evaluate ACC and MF1 result on BreaKHis dataset. The larger the better. \textbf{Bold} number means the best. The {\color{red!50}red} and {\color{green!50}green} boxes respectively represent the single model federated learning and personalized federated learning methods, and their individual clients use the unified model settings (ResNet17). The {\color{blue!50}blue} boxes represent the method of using heterogeneous models. The four client models are set to ResNet$\lbrace 17,11,8,5 \rbrace$, respectively. MH-pFLID achieves the best performance.}
  \resizebox{0.69\linewidth}{!}{
    \begin{tabular}{c|cc|cc|cc|cc|cc}
    \hline
    \multicolumn{1}{c|}{\multirow{2}[1]{*}{Method}} & \multicolumn{2}{c|}{high-resolution} & \multicolumn{2}{c|}{x2↓} & \multicolumn{2}{c|}{x4↓} & \multicolumn{2}{c|}{x8↓} & \multicolumn{2}{c}{Average} \\
     \cline{2-11} 
    \multicolumn{1}{c|}{} & ACC$\uparrow$   & MF1$\uparrow$   & ACC$\uparrow$   & MF1$\uparrow$   & ACC$\uparrow$  & MF1$\uparrow$   & ACC$\uparrow$   & MF1$\uparrow$   & ACC$\uparrow$   & MF1$\uparrow$ \\
    \hline
   \rowcolor{blue!10} Only Local Training & 0.7891  & 0.7319  & 0.8027  & 0.7461  & 0.7538  & 0.6852  & 0.6956  & 0.5867  & 0.7603  & 0.6875  \\
   \hline
   \rowcolor{red!10} FedAvg & 0.7406  & 0.6425  & 0.7908  & 0.7405  & 0.6892  & 0.6031  & 0.5774  & 0.4681  & 0.6995  & 0.6136  \\
   \rowcolor{red!10} FedAvg+FT & 0.7749 & 0.7218 & 0.8124 & 0.7511 & 0.7327 & 0.6628 & 0.6234 & 0.5073 & 0.7359  & 0.6608 \\
   \rowcolor{red!10} SCAFFOLD &0.7442  & 0.6512  & 0.8097  & 0.7533  & 0.6725  & 0.5963  & 0.5866  & 0.4732  & 0.7033  & 0.6185  \\
   \rowcolor{red!10} SCAFFOLD+FT & 0.7761 & 0.7229 & 0.8237 & 0.7709 & 0.7523 & 0.6872 & 0.6142 & 0.5005 & 0.7416  & 0.6704 \\
   \rowcolor{red!10} FedProx & 0.7354  & 0.6386  & 0.7873  & 0.7421  & 0.6944  & 0.6107  & 0.5821  & 0.4687  & 0.6998  & 0.6150 \\
   \rowcolor{red!10} FedProx+FT &0.7827 & 0.732 & 0.8055 & 0.7549 & 0.7548 & 0.6811 & 0.6071 & 0.4829 & 0.7375  & 0.6627 \\
    \hline
   \rowcolor{green!10} Ditto & 0.7304  & 0.6221  & 0.7661  & 0.6482  & 0.6065  & 0.5022  & 0.5931  & 0.4741  & 0.6740  & 0.5617  \\
   \rowcolor{green!10} APFL & 0.7444  & 0.6568  & 0.7992  & 0.7355  & 0.6227  & 0.5229  & 0.6133  & 0.4986  & 0.6949  & 0.6035  \\
   \rowcolor{green!10} FedRep & 0.7991  & 0.7618  & 0.8229  & 0.7697  & 0.7762  & 0.7182  & 0.6328  & 0.5091  & 0.7578  & 0.6897 \\
   \rowcolor{green!10} LG-FedAvg & 0.7972  & 0.7523  & 0.5655  & 0.4397  & 0.6131  & 0.5080  & 0.6080  & 0.4902  & 0.6460  & 0.5476  \\
    \hline
   \rowcolor{blue!10} FedMD & 0.7599  & 0.7083  & 0.8321  & 0.7829  & 0.7721  & 0.6893  & 0.6495  & 0.5439  & 0.7534  & 0.6811  \\
   \rowcolor{blue!10} FedDF & 0.7661  & 0.7253  & 0.8132  & 0.7629  & 0.7826  & 0.7342  & 0.6627  & 0.5627  & 0.7562  & 0.6963    \\
   \rowcolor{blue!10} pFedDF & 0.8233  & 0.7941  & 0.8369  & 0.7965  & 0.8121  & 0.7534  & 0.6843  & 0.6022  & 0.7892  & 0.7366  \\
   \rowcolor{blue!10} DS-pFL & 0.7842  & 0.7609  & 0.8334  & 0.7967  & 0.7782  & 0.7258  & 0.6327  & 0.5229  & 0.7571  & 0.7016 \\
   \rowcolor{blue!10}  KT-pFL & 0.8424  & 0.8133  & 0.8441  & 0.8011  & 0.7801  & 0.7325  & 0.7032  & 0.6219  & 0.7925  & 0.7422  \\
    \hline
    \rowcolor{blue!10} \model{} (Ours)  & \textbf{0.8929}  & \textbf{0.8658}  & \textbf{0.8992}  & \textbf{0.8787}  & \textbf{0.8661}  & \textbf{0.8327}  & \textbf{0.7751}  & \textbf{0.7130}  & \textbf{0.8583}  & \textbf{0.8226}  \\
    \hline
    \end{tabular}%
  \label{table1_rc}%
  }
\end{table*}%

\begin{table*}[ht]
  \centering
  \caption{The results of Image Classification Task with Different Label Distributions. This task includes breast cancer classification and OCT disease classification. We evaluate ACC and MF1 result in this task. The larger the better. \textbf{Bold} number means the best. MH-pFLID has the best performance.}
  \resizebox{1\linewidth}{!}{
    \begin{tabular}{c|cc|cc|cc|cc|cc|cc|cc|cc|cc}
    \hline
    \multicolumn{19}{c}{Breast cancer classification} \\
    \hline
    \multirow{2}[1]{*}{Method} & \multicolumn{2}{c}{ResNet} & \multicolumn{2}{c}{shufflenetv2} & \multicolumn{2}{c}{ResNeXt} & \multicolumn{2}{c}{squeezeNet} & \multicolumn{2}{c}{SENet} & \multicolumn{2}{c}{MobileNet} & \multicolumn{2}{c}{DenseNet} & \multicolumn{2}{c}{VGG} & \multicolumn{2}{c}{Average} \\
\cline{2-19}          & ACC$\uparrow$   & MF1$\uparrow$   & ACC$\uparrow$   & MF1$\uparrow$   & ACC$\uparrow$   & MF1 $\uparrow$  & ACC$\uparrow$   & MF1$\uparrow$   & ACC$\uparrow$   & MF1$\uparrow$   & ACC$\uparrow$   & MF1$\uparrow$   & ACC$\uparrow$   & MF1$\uparrow$   & ACC$\uparrow$   & MF1$\uparrow$   & ACC$\uparrow$   & MF1$\uparrow$ \\
    \hline
    Only Local Training & 0.59  & 0.455 & 0.845 & 0.8412 & 0.665 & 0.5519 & 0.84  & 0.7919 & 0.875 & 0.849 & 0.755 & 0.5752 & 0.855 & 0.6884 & 0.875 & 0.8515 & 0.7875  & 0.7005  \\
    \hline
    FedMD & 0.692 & 0.5721 & 0.823 & 0.8027 & 0.704 & 0.6087 & 0.875 & 0.8544 & 0.907 & 0.8745 & 0.762 & 0.6627 & 0.835 & 0.6493 & 0.842 & 0.8001 & 0.8050  & 0.7281  \\
    FedDF & 0.721 & 0.5949 & 0.817 & 0.8094 & 0.723 & 0.6221 & 0.893 & 0.8735 & 0.935 & 0.9021 & 0.757 & 0.6609 & 0.847 & 0.6819 & 0.833 & 0.7826 & 0.8158  & 0.7409  \\
    pFedDF & 0.755 & 0.6536 & 0.853 & 0.8256 & 0.741 & 0.6237 & 0.894 & 0.8742 & 0.935 & 0.9021 & 0.796 & 0.7219 & 0.879 & 0.7095 & 0.874 & 0.8521 & 0.8409  & 0.7703  \\
    DS-pFL & 0.715 & 0.6099 & 0.792 & 0.7734 & 0.765 & 0.6547 & 0.899 & 0.8792 & 0.935 & 0.9021 & 0.794 & 0.7331 & 0.853 & 0.6691 & 0.851 & 0.8266 & 0.8255  & 0.7560  \\
    KT-pFL & 0.765 & 0.6733 & 0.87  & 0.8331 & 0.755 & 0.6432 & 0.885 & 0.8621 & 0.935 & 0.9021 & 0.78  & 0.6931 & 0.865 & 0.6819 & 0.905 & 0.9023 & 0.8450  & 0.7739  \\
    \hline
    \model{} (Ours)  & \textbf{0.820}  & \textbf{0.6927} & \textbf{0.945}  & \textbf{0.9394} & \textbf{0.81}  & \textbf{0.7604} & \textbf{0.965} & \textbf{0.9457} & \textbf{0.982}  & \textbf{0.9709} & \textbf{0.815} & \textbf{0.7755} & \textbf{0.905} & \textbf{0.7287} & \textbf{0.974}  & \textbf{0.9583} & \textbf{0.9006}  & \textbf{0.8465}  \\
    \hline
    \multicolumn{19}{c}{OCT disease classification } \\
    \hline
    \multirow{2}[1]{*}{Method} & \multicolumn{2}{c}{ResNet} & \multicolumn{2}{c}{shufflenetv2} & \multicolumn{2}{c}{ResNeXt} & \multicolumn{2}{c}{squeezeNet} & \multicolumn{2}{c}{SENet} & \multicolumn{2}{c}{MobileNet} & \multicolumn{2}{c}{DenseNet} & \multicolumn{2}{c}{VGG} & \multicolumn{2}{c}{Average} \\
\cline{2-19}          & ACC$\uparrow$   & MF1$\uparrow$   & ACC$\uparrow$   & MF1$\uparrow$   & ACC$\uparrow$   & MF1$\uparrow$   & ACC$\uparrow$   & MF1$\uparrow$   & ACC $\uparrow$  & MF1$\uparrow$   & ACC$\uparrow$   & MF1$\uparrow$   & ACC$\uparrow$   & MF1$\uparrow$   & ACC$\uparrow$   & MF1$\uparrow$   & ACC$\uparrow$   & MF1$\uparrow$ \\
    \hline
    Only Local Training & 0.9162 & 0.9099 & 0.8922 & 0.8918 & 0.8694 & 0.8253 & 0.8472 & 0.8361 & 0.9388 & 0.9311 & 0.914 & 0.7236 & 0.9054 & 0.9   & 0.9262 & 0.9077 & 0.9012  & 0.8657  \\
    \hline
    FedMD & 0.8828 & 0.8349 & 0.8856 & 0.8531 & 0.8246 & 0.7822 & 0.8254 & 0.8021 & 0.8552 & 0.8321 & 0.9254 & 0.7542 & 0.9254 & 0.9119 & 0.9552 & 0.9293 & 0.8850  & 0.8375  \\
    FedDF & 0.854 & 0.8229 & 0.913 & 0.8936 & 0.865 & 0.8241 & 0.8054 & 0.7749 & 0.8926 & 0.8733 & 0.9178 & 0.7361 & 0.8958 & 0.8831 & 0.963 & 0.9308 & 0.8883  & 0.8424  \\
    pFedDF & 0.9364 & 0.9152 & 0.92  & 0.913 & 0.881 & 0.8327 & 0.863 & 0.8239 & 0.941 & 0.8952 & 0.931 & 0.7249 & 0.897 & 0.8829 & 0.961 & 0.9234 & 0.9163  & 0.8639  \\
    DS-pFL & 0.8432 & 0.8079 & 0.864 & 0.8604 & 0.874 & 0.8356 & 0.835 & 0.7449 & 0.8874 & 0.8821 & 0.8998 & 0.7532 & 0.8592 & 0.8264 & 0.8814 & 0.8731 & 0.8680  & 0.8230  \\
    KT-pFL & 0.9532 & 0.9392 & 0.965 & 0.963 & 0.8594 & 0.8466 & 0.9136 & 0.9067 & 0.955 & 0.943 & 0.9622 & 0.8099 & 0.9038 & 0.8794 & 0.927 & 0.9022 & 0.9299  & 0.8988  \\
    \hline
    \model{} (Ours)  & \textbf{0.9644} & \textbf{0.9516} & \textbf{0.992} & \textbf{0.983} & \textbf{0.898} & \textbf{0.8646} & \textbf{0.9742} & \textbf{0.966} & \textbf{0.971} & \textbf{0.9661} & \textbf{0.9544} & \textbf{0.8162} & \textbf{0.9162} & \textbf{0.9121} & \textbf{0.966} & \textbf{0.9511} & \textbf{0.9545}  & \textbf{0.9263}  \\
    \hline
    \end{tabular}%
  \label{table_lc}%
  }
\end{table*}%

\section{Experiment Setup}
\subsection{Tasks and Datasets}

We verify the effectiveness of \model{} on 4 non-IID tasks.
%, namely: different resolution medical image classification task, medical image classification task with different label distributions, Time-series classification task and medical image segmentation task.

\textbf{A. Medical image classification (different resolution).} We use the Breast Cancer Histopathological Image Database (BreaKHis) \cite{7312934}. We perform x2↓, x4↓, and x8↓ downsampling on the high-resolution images \cite{7312934}. Each resolution of medical images is treated as a client, resulting in four clients in total. The dataset for each client was randomly divided into training and testing sets at a ratio of 7:3, following previous work. For the same image with different resolutions, they will be used in either the training set or the testing set. In this task, we employed ResNet$\lbrace 17,11,8,5 \rbrace$.

\textbf{B. Medical image classification (different label distributions).} This task includes a breast cancer classification task and an OCT disease classification task. We design eight clients, each corresponding to a distinct heterogeneous model. These models include ResNet \cite{he2015deep}, ShuffleNetV2 \cite{ma2018shufflenet}, ResNeXt \cite{xie2017aggregated}, SqueezeNet \cite{iandola2016squeezenet}, SENet \cite{hu2018squeeze}, MobileNetV2 \cite{sandler2018mobilenetv2}, DenseNet \cite{huang2017densely}, and VGG \cite{simonyan2014very}. Similar to FedAvg, we apply non-IID label distribution methods to BreaKHis (RGB images) and OCT2017 (grayscale images) \cite{kermany2018identifying} across 8 clients.Specifically, in different clients, the label of each client is set to be different. Besides, the data distribution is also different among clients.

\textbf{C. Medical time-series classification.} We used the Sleep-EDF dataset 
 \cite{goldberger2000physiobank} for the classification task of time series under non-IID distribution. We designed three clients using the TCN \cite{bai2018empirical}, Transformer \cite{2021A} and RNN \cite{xie2024trls}.

\textbf{D. Medical image segmentation.} Here, we focus on polyp segmentation \cite{dong2021polyp}. The dataset consists of endoscopic images collected and annotated from four centers, with each center's dataset treats as a separate client. Each client utilized a specific model, including Unet++ \cite{zhou2019unet++}, FCN \cite{long2015fully}, Unet \cite{ronneberger2015u}, and Res-Unet \cite{diakogiannis2020resunet}.

%different learning rates for the two training stages, with  including the \model{} framework, 

\subsection{Implementation Details}

\model{} adopts learning rate of 0.0001 and 0.00001 for the knowledge injection and knowledge distillation stage. The batch size is set to 8. In experiments, all frameworks have a communication round of 100. Local training epochs are 5 (4 epochs in the first stage and 1 round in the second stage for \model{}). For classification, $\mathcal{L}^l_{\textit{inj}}$, $\mathcal{L}^m_{\textit{inj}}$ and $\mathcal{L}^m_{\textit{dis}}$ are cross-entropy loss. $\mathcal{L}^{con}_{\textit{dis}}$ is KL loss \cite{aggarwal2021generative}. And for segmentation tasks, $\mathcal{L}^l_{\textit{inj}}$, $\mathcal{L}^m_{\textit{inj}}$ and $\mathcal{L}^m_{\textit{dis}}$ are Dice loss. $\mathcal{L}^{con}_{\textit{dis}}$ still is KL loss. $\lambda^l_{\textit{inj}}$ and $\lambda^l_{\textit{dis}}$ are set to 0.9. $\lambda^m_{\textit{inj}}$ and $\lambda^{con}_{\textit{dis}}$ are 0.1.The performance evaluation of the classification task is based on two metrics, accuracy (ACC) and macro-averaged F1-score (MF1), providing a comprehensive assessment of the model's robustness. Additionally, Dice is used to evaluate the segmentation task performance across frameworks. 

%Moreover, we implement \model{} using PyTorch 1.10 \cite{paszke2019pytorch} and train it on an NVIDIA GeForce RTX 3090 Ti GPU. We have included more baseline, datasets, training settings and model structure details in the supplementary materials.

% Table generated by Excel2LaTeX from sheet '时间序列'
\begin{table}[t]
  \centering
  \caption{The results of the time-series classification task. \model{} performs the best. We evaluate ACC and MF1 result in this task. The larger the better. Bold number means the best. MH-pFLID has the best performance.}
  \resizebox{1\linewidth}{!}{
    \begin{tabular}{c|cc|cc|cc|cc}
    \hline
    \multirow{2}[1]{*}{Method} & \multicolumn{2}{c}{TCN} & \multicolumn{2}{c}{Transformer} & \multicolumn{2}{c|}{RNN} & \multicolumn{2}{c}{Average} \\
\cline{2-9}          & ACC$\uparrow$   & MF1$\uparrow$   & ACC$\uparrow$   & MF1$\uparrow$   & ACC$\uparrow$   & MF1$\uparrow$   & ACC$\uparrow$   & MF1$\uparrow$ \\
    \hline
    Only Local Training & 0.9073 & 0.8757 & 0.8053 & 0.8001 & 0.8012 & 0.7263 & 0.8379  & 0.8007  \\
    \hline
    FedMD & 0.9334 & 0.9225 & 0.7934 & 0.7966 & 0.793 & 0.7072 & 0.8399  & 0.8088  \\
    FedDF & 0.9146 & 0.8893 & 0.7988 & 0.8042 & 0.7881 & 0.6855 & 0.8338  & 0.7930  \\
    pFedDF & 0.9173 & 0.8957 & 0.827 & 0.8309 & 0.8137 & 0.7713 & 0.8527  & 0.8326  \\
    DS-pFL & 0.9133 & 0.9033 & \multicolumn{1}{r}{0.8253} & \multicolumn{1}{r|}{0.8301} & 0.8042 & 0.7539 & 0.8476  & 0.8291  \\
    KT-pFL & 0.9240 & 0.9089 & 0.8419 & 0.8466 & 0.8204 & 0.7722 & 0.8621  & 0.8426  \\
    \hline
    \model{} (Ours)  & \textbf{0.9439} & \textbf{0.9248} & \textbf{0.8725} & \textbf{0.876} & \textbf{0.824} & \textbf{0.7773} & \textbf{0.8801}  & \textbf{0.8594}  \\
    \hline
    \end{tabular}%
  \label{tab_tc}%
  % \vspace{-0.6cm}
  }
\end{table}%

% Table generated by Excel2LaTeX from sheet '分割'
\begin{table}[t]
  \centering
  \caption{ %The results of medical image segmentation. The evaluation indicator used is Dice.
  For the medical image segmentation task, we evaluate the Dice result on Polyp dataset. The larger the better. \textbf{Bold} number means the best.
  The {\color{red!50}red} and {\color{green!50}green} boxes represent the single model federated learning and personalized federated learning methods, respectively. Their individual clients use the unified model settings (Unet). The {\color{blue!50}blue} boxes represent the method of using heterogeneous models in each client. The four client models are set to Unet++, FCN, Unet, and ResUnet, respectively. MH-pFLID achieves the best segmentation results.}
   \resizebox{0.9\linewidth}{!}{
    \begin{tabular}{c|c|c|c|c|c}
    \hline
    Methods & Client1 & Client2 & Client3 & Client4 & Average \\
    \hline
    \rowcolor{red!10} {FedAvg} & 0.5249  & 0.4205  & 0.5676  & 0.5500  & 0.5158  \\
   \rowcolor{red!10} FedAvg+FT & 0.6047  & 0.4762  & 0.7513  & 0.6681  & 0.6251  \\
    \rowcolor{red!10} SCAFFOLD & 0.5244  & 0.3591  & 0.5935  & 0.5713  & 0.5121  \\
    \rowcolor{red!10} SCAFFOLD+FT & 0.5937  & 0.4312  & 0.8231  & 0.7208  & 0.6422  \\
    \rowcolor{red!10} FedProx & 0.5529  & 0.4674  & 0.5403  & 0.6301  & 0.5477  \\
    \rowcolor{red!10} FedProx+FT & 0.7441  & 0.5701  & 0.7438  & 0.6402  & 0.6746  \\
    \hline
   \rowcolor{green!10} Ditto & 0.5720  & 0.4644  & 0.6648  & 0.6416  & 0.5857  \\
    \rowcolor{green!10} APFL  & 0.6120  & 0.5095  & 0.6333  & 0.5892  & 0.5860  \\
    \rowcolor{green!10} LG-FedAvg & 0.6053  & 0.5062  & 0.7371  & 0.5596  & 0.6021  \\
    \rowcolor{green!10} FedRep & 0.5809  & 0.3106  & 0.7088  & 0.7023  & 0.5757  \\
    \hline
    \rowcolor{green!10} FedSM & 0.6894  & 0.6278  & 0.8021  & 0.7391  & 0.7146  \\
    \rowcolor{green!10} LC-Fed & 0.6233  & 0.4982  & 0.8217  & 0.7654  & 0.6772  \\
    \hline
    \rowcolor{blue!10} Only Local Training & 0.7049  & 0.4906  & 0.8079  & 0.7555  & 0.6897  \\
    \hline
    \rowcolor{blue!10} \model{} (Ours)  & \textbf{0.7591}  & \textbf{0.6528}  & \textbf{0.8543}  & \textbf{0.7752}  & \textbf{0.7604}  \\
    \hline
    \end{tabular}%
  \label{tab_seg}%
  }
  %\vspace{-0.2in}
\end{table}%

\begin{figure*}[t]
\vskip 0.2in
\includegraphics[width=0.92\textwidth]{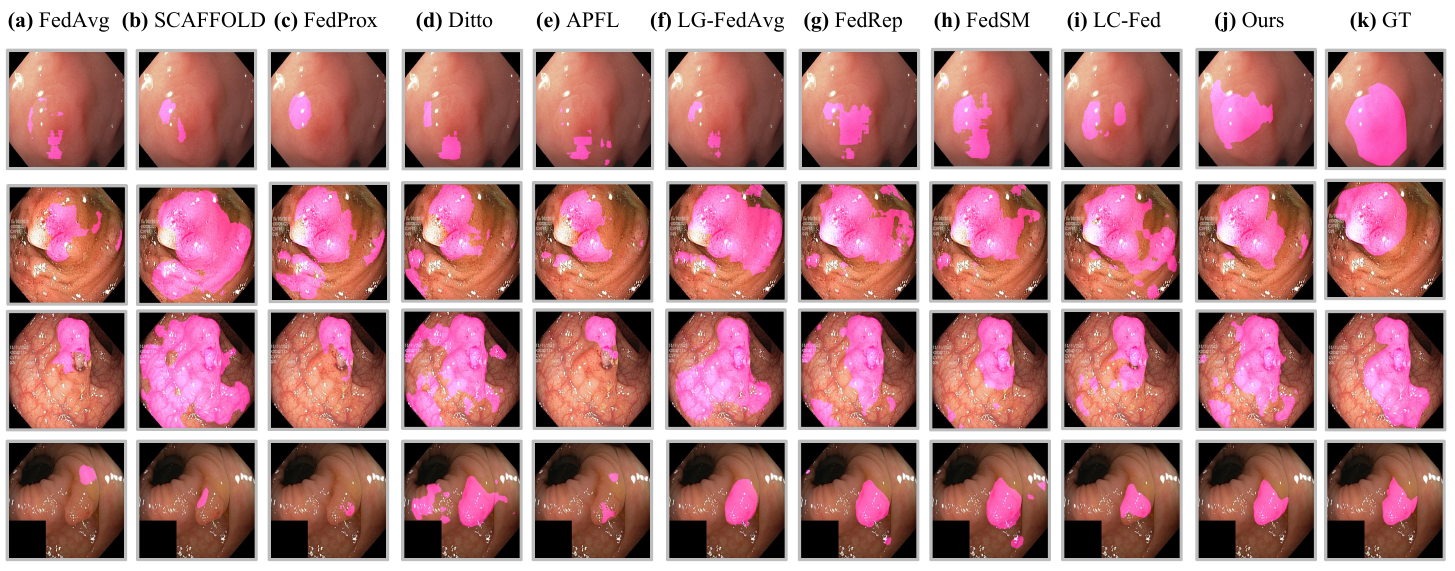}
\centering
\caption{Visualized comparison of Federated Learning in medical image segmentation. We randomly select four samples from different clients to form the visualization. (a-j)  Segmentation results by a model trained with FedAVG, SCAFFOLD, FedProx, Ditto, APFL, LG-FedAvg, FedRep, FedSM, LC-Fed, and our method \model{}; (k) Ground truths (denoted as ‘GT’).}
\label{fig_4}
\vskip -0.2in
\end{figure*}

\begin{table}[t]
  \centering
  \caption{GFLOPS and parameters of local heterogeneous models and messenger models in various tasks. The smaller the better. Bold number means the best. Among the four tasks, the GFLOPS and parameters of the messenger models are much smaller than those of the local models.}
  \resizebox{0.95\linewidth}{!}{
    \begin{tabular}{c|c|ccc}
    \toprule
    \textbf{Tasks} & \textbf{Dataset} & \textbf{Model} & \textbf{GFLOPS} & \textbf{\#Params} \\
    \hline
    \multicolumn{1}{c|}{\multirow{5}[4]{*}{\makecell{Medical Image \\ Classification \\ (Different Resolution)}}} & \multicolumn{1}{c|}{\multirow{5}[4]{*}{\makecell{BreaKHis\\(384x384x3\\- 48x48x3)}}} & ResNet17 & 3.495  & 4.231M \\
          &       & ResNet11 & 0.667  & 2.104M \\
          &       & ResNet8 & 0.140  & 1.558M \\
          &       & ResNet5 & 0.044  & 1.359M \\
\cline{3-5}          &       & Messeger & \textbf{0.01-0.07} & \textbf{0.035M} \\
    \hline
    \multicolumn{1}{c|}{\multirow{18}[8]{*}{\makecell{Medical Image \\ Classification \\ (Different Label \\ Distributions)}}} & \multicolumn{1}{c|}{\multirow{9}[4]{*}{\makecell{BreaKHis\\(384x384x3)}}} & ResNet & 10.020  & 11.111M \\
          &       & Shufflenetv2 & 1.719  & 1.730M \\
          &       & ResNeXt & 41.245  & 7.930M \\
          &       & squeezeNet & 7.774  & 1.832M \\
          &       & SENet & 80.370  & 12.372M \\
          &       & MobileNet & 1.870  & 1.934M \\
          &       & DenseNet & 13.461  & 1.147M \\
          &       & VGG   & 57.524  & 40.045M \\
\cline{3-5}          &       & Messeger & \textbf{0.070} & \textbf{0.032M} \\
\cline{2-5}          & \multicolumn{1}{c|}{\multirow{9}[4]{*}{\makecell{OCT 2017\\(256x256x1)}}} & ResNet & 4.351  & 11.090M \\
          &       & Shufflenetv2 & 0.735  & 1.712M \\
          &       & ResNeXt & 18.256  & 7.910M \\
          &       & squeezeNet & 3.342  & 1.820M \\
          &       & SENet & 35.644  & 12.363M \\
          &       & MobileNet & 0.812  & 1.921M \\
          &       & DenseNet & 5.954  & 1.14M \\
          &       & VGG   & 25.501  & 40.020M \\
\cline{3-5}          &       & Messeger  & \textbf{0.012 } & \textbf{0.035M} \\
    \hline
    \multirow{4}[4]{*}{\makecell{ Medical Time-series \\ Classification Task}} & \multirow{4}[4]{*}{\makecell{Sleep-EDF\\(1x3000)}} & TCN   & 17.101  & 1.182M \\
          &       & Transformer & 1.224  & 1.134M \\
          &       & RNN   & 5.200  & 1.236M \\
\cline{3-5}          &       & Messeger & \textbf{0.411} & \textbf{0.003M} \\
    \hline
    \multicolumn{1}{c|}{\multirow{5}[4]{*}{\makecell{Medical Image \\ Segmentation Task}}} & \multicolumn{1}{c|}{\multirow{5}[4]{*}{\makecell{Polyp\\(256x256x3)}}} & Unet++ & 34.906  & 10.421M \\
          &       & FCN   & 54.742  & 32.560M \\
          &       & Unet  & 56.435  & 33.090M \\
          &       & ResUnet & 25.572  & 19.913M \\
\cline{3-5}          &       & Messeger  & \textbf{0.681} & \textbf{0.196M} \\
    \bottomrule
    \end{tabular}%
    }
  \label{tab_FP}%
  \vspace{-0.4cm}
\end{table}%

\section{Results}

%\subsection{The results of the classification task with medical images at different resolutions.}. For example, high-resolution images are trained using ResNet17, and x8↓ images are trained using ResNet5. compared to federated learning architectures where each client uses a homogeneous model and existing federated learning frameworks where each client uses heterogeneous models

\subsection{Medical Image Classification (Different Resolutions)} 
In this task, we employ models from the ResNet family to train breast cancer medical images at different resolutions. For low-resolution images, we use shallow ResNet models for training, while for high-resolution images, we employ deeper and more complex ResNet models. In \cref{table1_rc}, compared to other federated learning frameworks, \model{} achieves the best performance on \textit{all} clients of different resolutions, including the original high-resolution, half(``$\times 2\downarrow$''), quarter (``$\times 4\downarrow$''), and one eighth (``$\times 8\downarrow$''). This indicates that \model{}, based on the injection and distillation paradigm, effectively enables local heterogeneous models within the same family to learn global knowledge, thereby enhancing the performance of local models. Furthermore, \model{} demonstrates a more significant advantage in terms of the MF1 metric, highlighting its ability to improve the robustness of local heterogeneous models.

%\subsection{The results of the medical image classification task with different label distributions}

\subsection{Medical Image Classification (Different Label Distributions)} 
In \cref{table_lc}, the experimental results for the medical image classification task with different label distributions, where each client uses heterogeneous models, show that \model{} achieves the optimal results. This demonstrates that, compared to heterogeneous federated learning methods based on soft predictions, the Injection and Distillation approach of \model{} has advantages. It can more effectively utilize knowledge from other clients to guide local client learning. Compared to local training alone, \model{} enhances the local performance of each heterogeneous model. This indicates that our proposed feature adaptation method, aligning global and local features, effectively alleviates the issue of client shift when guiding client training for each heterogeneous model.

%\subsection{The results of the Time-series classification task}
\subsection{Time-series Classification}
The experimental results in \cref{tab_tc} show that \model{} achieves the best results under different types of neural networks. This further demonstrates the superiority of \model{} in federated learning of heterogeneous models.

%\subsection{The results of the image segmentation task}

\subsection{Medical Image Segmentation} 
We have once again validated the effectiveness of \model{} in medical image segmentation tasks. \cref{tab_seg} presents the results of federated learning in the segmentation task, demonstrating that \model{} achieves the best experimental outcomes. This indicates \model{} applicability across multiple tasks. The experimental results further highlight that \model{} effectively enhances the local model performance for each client in various tasks, surpassing existing personalized approaches for homogeneous models. Meanwhile, the visualization results in \cref{fig_4} show that the segmentation results of \model{} are closer to ground truth.

%\subsection{ablation experiments}

\subsection{GFLOPS and Parameters}
We compare the GFLOPS and parameter of the messenger model with local heterogeneous models in four tasks. The results of \cref{tab_FP} show that the GFLOPS and parameters of the messenger model are much smaller than those of the local heterogeneous model.

\begin{table}[ht]
  \centering
  \caption{The ablation experiments of \model{}. We remove some essential modules to verify the effectiveness of each module. We perform experiments on breast cancer classification (different label distributions) and medical image segmentation tasks. We observe that though those experiments outperform only local training, they suffer different levels of performance decrease.}
  \resizebox{1\linewidth}{!}{
    \begin{tabular}{c|cc|c}
    \hline
    \multirow{2}[1]{*}{Methods} & \multicolumn{2}{c|}{Breast Cancer} & Segmentation \\
\cline{2-4}          & ACC$\uparrow$   & MF1$\uparrow$   & Dice$\uparrow$ \\
    \hline
    \model{} & \textbf{0.9006} & \textbf{0.8465} & \textbf{0.7604} \\
    \hline
    w/o Messenger Head & 0.8657 & 0.7921 & 0.7391 \\
    w/o Messenger Body  & 0.8432 & 0.7709 & 0.7294 \\
    w/o Information Receiver & 0.8631 & 0.8021 & 0.7339 \\
    w/o Information Transmitter & 0.8791 & 0.8249 & 0.7405 \\
    w/o Information Receiver \& Transmitter & 0.8479 & 0.7853 & 0.7306 \\
    \hline
    Only Local Training & 0.7875 & 0.7005 & 0.6897 \\
    \hline
    \end{tabular}%
  \label{table_ab}%
  }
\end{table}%

\subsection{Ablation Studies}
To verify the effectiveness of the proposed components in \model{}, a comparison between \model{} and its four components on breast cancer classification in different label
distributions and segmentation tasks is given in \cref{table_ab}. The four components are as follows: (1) w/o messenger head and w/o messenger body: the distilled head or body does not participate in the global aggregation stage. (2) w/o information receiver or information transmitter indicates that the information receiver or information transmitter (IT) are replaced with the feature add operation. The experimental results indicate that more parameter sharing is beneficial for \model{}. information receiver and information transmitter operations effectively improve the performance of local heterogeneous models. 

% Table generated by Excel2LaTeX from sheet 'Sheet2'
\begin{table*}[htbp]
  \centering
  \caption{Generalizability experiments of the messenger
model. Our method increases the performance not only on the models' own clients, but on other untrained clients.}
  \resizebox{1\linewidth}{!}{
  
    \begin{tabular}{c|ccccc|c|ccccr|c|cccc}
     \hline
          & \multicolumn{4}{c}{Only Local Training} &       &       & \multicolumn{4}{c}{Ours}      &       &       & \multicolumn{4}{c}{Improvement} \\
     \hline
          & Client1 & Client2 & Client3 & Client4 &       &       & Client1 & Client2 & Client3 & Client4 &       &       & Client1 & Client2 & Client3 & Client4 \\
    Unet++ & 0.7049 & 0.0586 & 0.3016 & 0.2239 &       & Unet++ & 0.7591 & 0.1596 & 0.3542 & 0.4497 &       & Unet++ & {+7.1\%} & {+63.3\%} & {+14.9\%} & {+50.2\%} \\
    Unet  & 0.1393 & 0.4906 & 0.2231 & 0.404 &       & Unet  & 0.1418 & 0.6528 & 0.225 & 0.4654 &       & Unet  & {+0.3\%} & {+16.2\%} & {+0.2\%} & {+6.1\%} \\
    ResUnet & 0.2433 & 0.1241 & 0.8079 & 0.4527 &       & ResUnet & 0.3239 & 0.2929 & 0.8543 & 0.6066 &       & ResUnet & {+8.1\%} & {+16.9\%} & {+4.6\%} & {+15.4\%} \\
    FCN   & 0.442 & 0.3247 & 0.4597 & 0.7555 &       & FCN   & 0.3403 & 0.434 & 0.4855 & 0.7752 &       & FCN   & {-10.2\%} & {+10.9\%} & {+2.6\%} & {+2.0\%} \\
    \hline
    \end{tabular}%
    }
    
  \label{tab:general}%
  \vspace{-0.2cm}
\end{table*}% 

\begin{figure}[t]
\includegraphics[width=0.40\textwidth]{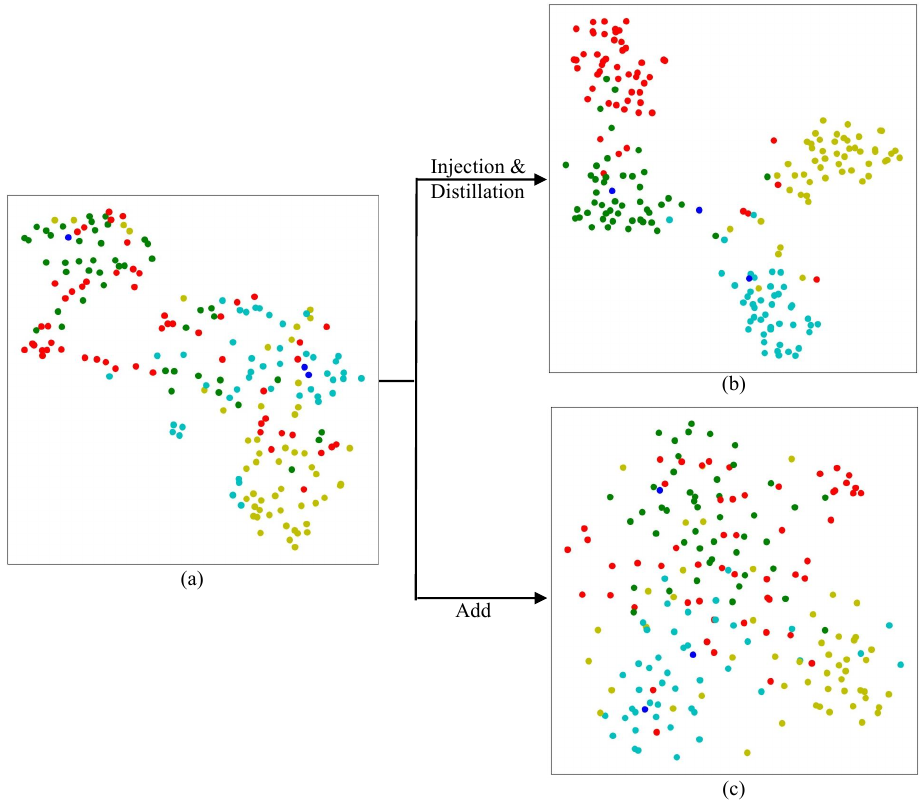}
\centering
\caption{t-SNE map of the 7th client (DenseNet) through injection \& distillation or add under breast cancer classification task (different label distributions). Different colored dots represent different categories. 
%The add operation adds the features extracted from the messenger body to the features extracted from the local body. 
(a-c) are the t-SNE map of (a) the features extracted from the local model body, (b) local model body features after injection \& distillation, and (c) the same feature by replacing the injection \& distillation process with a simple \textit{add} operation and retrained. The experimental results show that the features generated by injection \& distillation are more distinguishable compared to the straightforward \textit{add} design.  }
\label{fig_5}
\vspace{-0.6cm}
\end{figure}

\subsection{Feature Distribution t-SNE of Injection \& Distillation Operations} 
We train the local heterogeneous model for 20 epochs without any other operations (The feature distribution is shown in \cref{fig_5}(a)).  Next, we will perform 5 communication rounds of \textit{injection-distillation} or \textit{add} on the trained local heterogeneous model. The experimental results show that the \textit{injection-distillation} operation (\cref{fig_5}(b)) generates more discriminative features compared to the \textit{add} operation (\cref{fig_5}(c)). This is more conducive to subsequent tasks.

\subsection{Generalizability Experiments of the Messenger
Model}

We evaluate local models directly on other clients without training on those clients' data on the segmentation task in Table 7.
% The left part of the following \cref{tab:general} is only local training results, where we train each module independently on their own client.
% % (Client 1: Unet++, Client 2: Unet, Client 3: ResUnet, Client 4: FCN).
% The middle part shows the result of our model trained including messenger information. The right part is the percentage of improvement. 
We see performance increases not only on the models' own clients, but on other untrained clients. These results verify that our messenger would increase the generalizability of the local model, thus it successfully collects generalized information from other clients.

\subsection{The Disentanglement of the Bases from the Messenger}
 To evaluate the disentanglement, we calculate how orthogonal are those bases by calculating $e=\|BB^{\top}-I\|_f$ as the training proceeds, where $B$ is the bases matrix, $I$ is the identity matrix, and $\|\cdot\|_f$ is the Frobenius norm. 
%This equation means we calculate the inner product of all base pairs, and i
If they are fully orthogonal to each other, $\|BB^{\top}-I\|_f$ should be equal to 0. Our experiment in the following \cref{tab:disen} shows that, as the training proceeds, $e$ clearly drops, meaning our transmitter and receiver give good disentanglements.
%Even without the explicit disentangle loss, our transmitter and receiver would still give good disentanglement for the bases in the messenger output. 

\begin{table}[htbp]
  \centering
  \caption{The disentanglement of feature bases changes in training process.}
  \resizebox{0.4\linewidth}{!}{
  
    \begin{tabular}{cc}
    \toprule
    Rounds & $e =\|BB^{\top}-I\|_f$ \\
    \midrule
    1     & 0.44 \\
    5     & 0.37 \\
    10    & 0.11 \\
    20    & 0.10 \\
    \bottomrule
    \end{tabular}%
    
    }
  \label{tab:disen}%

\end{table}%

% \begin{table}
%   \centering
%   \vspace{-0.4cm}
%   \caption{\textcolor{blue}{Results on PROMISE 12 compared with SAT.}}
%    \resizebox{0.6\linewidth}{!}{
%    {\color{blue}
%     \begin{tabular}{c|ccc}
%       \hline
%           & SAT & Local(Unet) & Ours(Unet)\\
%      \hline      
%     DSC& 82.31 & 80.67 & 87.25 \\
%      \hline
%     \end{tabular}%
%     }
%     }
%   \label{tab:sat}%
%   \vspace{-0.6cm}
% \end{table}%

% {
% \color{blue}
% \textbf{Comparison with foundation models.} We compare the foundation model method MedSAM with our approach to Polyp segmentation tasks. The results are shown in \cref{tab:medsam}. We also compare the foundation-model-based method SAT on the Prostate segmentation task on PROMISE12 dataset (shown in \cref{tab:sat}). Our method outperforms both of them.
% }


\section{Limitations and Conclusion}

Our method has demonstrated its effectiveness in medical classification and segmentation tasks, yet we have not validated and refined our approach in medical object detection, image registration, medical 3D reconstruction, etc.. In future work, the potential of our method in these areas awaits further verification and enhancement. \model{} effectively addresses challenges faced by existing personalized federated learning approaches for heterogeneous models. These challenges include collecting and labeling public datasets, and computational burden on local clients and servers. \model{}, based on injection and distillation paradigm, offers a solution to these issues. \model{} introduces a lightweight messenger model in each client and designs information receiver and transmitter. These can enable local heterogeneous models to transfer information from other clients well under non-IID distribution. Extensive experiments demonstrate superiority of \model{} over existing frameworks for federated learning with heterogeneous models.

\section*{Acknowledgements}
This work was supported by the National Key R$\&$D Program of China under Grant No.2022YFB2703301.

The authors gratefully acknowledge the insightful and constructive discussions provided by Xiao Chen at United Imaging Intelligence.

\section*{Impact Statements}

This paper presents work whose goal is to advance the field of Machine Learning. There are many potential societal consequences of our work, none which we feel must be specifically highlighted here.

\nocite{langley00}

\bibliography{example_paper}
\bibliographystyle{icml2024}

%%%%%%%%%%%%%%%%%%%%%%%%%%%%%%%%%%%%%%%%%%%%%%%%%%%%%%%%%%%%%%%%%%%%%%%%%%%%%%%
%%%%%%%%%%%%%%%%%%%%%%%%%%%%%%%%%%%%%%%%%%%%%%%%%%%%%%%%%%%%%%%%%%%%%%%%%%%%%%%
% APPENDIX
%%%%%%%%%%%%%%%%%%%%%%%%%%%%%%%%%%%%%%%%%%%%%%%%%%%%%%%%%%%%%%%%%%%%%%%%%%%%%%%
%%%%%%%%%%%%%%%%%%%%%%%%%%%%%%%%%%%%%%%%%%%%%%%%%%%%%%%%%%%%%%%%%%%%%%%%%%%%%%%
\newpage
\appendix
\onecolumn
\section{Related Works Supplement}

In this section, we provide a detailed supplement to the comparison of personalized federated learning in this paper.

APFL \cite{apfl} adopts the joint prediction mode which is a mixture of global model and local model with adaptive weight. LG-FedAvg \cite{lg-fedavg} and FedRep \cite{fedrep} use parameter decoupling to jointly learn the global part and local part of the client model, while only the global part was sent to the server. The difference between them is the definition of which part of the model is the global part. kNN-Per \cite{knnper} uses the output of the K global representations closest to the global representation of the input to be evaluated to make predictions. Ditto \cite{li2021ditto} adds a regular term to the original local loss function of each client to measure the deviation between the local model and the global model.

\section{Baselines}

In the medical image classification task (different resolution), we selected FedAvg, SCAFFOLD, FedProx, and their fine-tuned methods \cite{NEURIPS2022_449590df} same as previous work \cite{fedrep}. Among the personalized Federated Learning methods, we compared FedRep, LG-FedAvg, APFL, and Ditto. For heterogeneous model federated learning, we chose FedMD, FedDF, pFedDF, DS-pFL, and KT-pFL.

In the medical image classification task (different label distrbutions), we compared various methods, including local training of clients with heterogeneous models and existing heterogeneous model federated learning approaches (FedMD, FedDF, pFedDF, DS-pFL, and KT-pFL). 

The baseline used in the medical time-series classification task is the same as the medical image classification task (different label distrbutions).

For image segmentation tasks, we compared various approaches, including local training of clients and a variety of personalized federated learning techniques, as well as methods for learning a single global model and their fine-tuned versions. Among the personalized methods, we also chose FedRep, LG-FedAvg, APFL, and Ditto. We simultaneously added LC-Fed \cite{lcfed} and FedSM \cite{fedsm} which are effective improvements for FedRep and APFL in the federated segmentation domain.

\section{Datasets}

\textbf{A. Medical image classification (different resolution).} We used the Breast Cancer Histopathological Image Database (BreaKHis) \cite{7312934}. We treat the original image as a high-resolution image. Then, the Bicubic downsampling method is used to downsample the high-resolution image, obtaining images with resolutions of x2 ↓, x4 ↓, and x8 ↓, respectively. Each resolution of medical images was treated as a separate client, resulting in four clients in total. Each client has the same number of images with consistent label distribution, but the image resolution is different for each client. The dataset for each client was randomly divided into training and testing sets at a ratio of 7:3, following previous work. In this task, we employed a family of models such as ResNet$\lbrace 17,11,8,5 \rbrace$.

\textbf{B. Medical image classification (different label distributions).} This task includes a breast cancer classification task and an OCT disease classification task. We designed eight clients, each corresponding to a distinct heterogeneous model. These models included ResNet \cite{he2015deep}, ShuffleNetV2 \cite{ma2018shufflenet}, ResNeXt \cite{xie2017aggregated}, SqueezeNet \cite{iandola2016squeezenet}, SENet \cite{hu2018squeeze}, MobileNetV2 \cite{sandler2018mobilenetv2}, DenseNet \cite{huang2017densely}, and VGG \cite{simonyan2014very}. Similar to the previous approach, we applied non-IID label distribution methods to the OCT2017 (grayscale images) \cite{kermany2018identifying} and BreaKHis (RGB images) across the 8 clients.

For the breast cancer classification task, we have filled in the data quantity to 8000 and allocated 1000 pieces of data to each client. The ratio of training set to testing set for each client is 8:2.

For the OCT disease classification task, we randomly selected 40000 pieces of data, with 5000 pieces per client. The ratio of training set to test set is also 8:2. 

\textbf{C. Medical time-series classification.} We used the Sleep-EDF dataset 
 \cite{goldberger2000physiobank} for the classification task of time series under Non-IID distribution. We divided the Sleep-EDF dataset evenly among three clients.  The ratio of training set to testing set for each client is 8:2. We designed three clients using the TCN \cite{bai2018empirical}, Transformer \cite{2021A} and RNN \cite{xie2024trls}. 

\textbf{D. Medical image segmentation.} Here, we focus on polyp segmentation \cite{dong2021polyp}. The dataset for this task consisted of endoscopic images collected and annotated from four different centers, with each center's dataset treated as a separate client. Thus, there were four clients in total for this task. The number of each client are 1000, 380, 196 and 612. The ratio of training set to testing set for each client is 1:1.
Each client utilized a specific model, including Unet++ \cite{zhou2019unet++}, FCN \cite{long2015fully}, Unet \cite{ronneberger2015u}, and Res-Unet \cite{diakogiannis2020resunet}.

\section{Training Settings}

\subsection{Evaluation Indicators}
The performance evaluation of the classification task is based on two metrics, accuracy (ACC) and macro-averaged F1-score (MF1), providing a comprehensive assessment of the model's robustness. Additionally, Dice is used to evaluate the segmentation task performance across frameworks. 

\textbf{A. Accuracy.} Accuracy is the ratio of the number of correct judgments to the total number of judgments. 

\textbf{B. Macro-averaged F1-score.} First, calculate the F1-score for each recognition category, and then calculate the overall average value. 

\textbf{C. Dice.} It is a set similarity metric commonly used to calculate the similarity between two samples, with a threshold of [0,1]. In medical images, it is often used for image segmentation, with the best segmentation result being 1 and the worst result being 0. The Dice coefficient calculation formula is as follows:
\begin{small} 
\begin{equation} 
Dice = \frac{{2*(pred \cap true)}}{{pred \cup true}}
\end{equation} 
\end{small}
Among them, $pred$ is the set of predicted values, $true$ is the set of groudtruth values. And the numerator is the intersection between $pred$ and $true$. Multiplying by 2 is due to the repeated calculation of common elements between $pred$ and $true$ in the denominator. The denominator is the union of $pred$ and $true$.

\subsection{Loss Function}
Many loss functions have been applied in this article, and here are some explanations for them.The cross entropy loss function is very common and will not be explained in detail here. 
We mainly explain Dice loss.% and KL Loss.

Dice Loss applied in the field of image segmentation. It is represented as:

\begin{small} 
\begin{equation} 
DiceLoss = 1- \frac{{2*(pred \cap true)}}{{pred \cup true}}
\end{equation} 
\end{small}

The Dice loss and Dice coefficient are the same thing, and their relationship is:

\begin{small} 
\begin{equation} 
DiceLoss = 1- Dice
\end{equation} 
\end{small}

In knowledge distillation, we use KL divergence loss to optimize the distribution difference between the messenger model and the local model. We try to constrain the output of the messenger head and local head model to be under the same distribution, so that the knowledge can be distillated from the local model to the messenger model. The KL loss is defined as:

\begin{small} 
\begin{equation} 
KLLoss = \sum_{i=0}^m[y_ilog(y_i)-\hat{y_i}log(\hat{y_i})]
\end{equation} 
\end{small}

where $m$ represents the total number of client data. $\hat{y_i}$ and $y_i$ represent the outputs of the messenger model and the local model, respectively.

\subsection{Public Datasets for other Federated Learning of Heterogeneous Models}

In this section, we mainly describe the setting of public datasets for methods such as FedMD, FedDF, DS-pFL and KT-pFL.

\textbf{A. Medical image classification (different resolution).} We select 100 pieces of data from each client and put them into the central server as public data, totaling 400 pieces of data as public data. In order to better obtain soft predictions for individual clients, the image resolution of the publicly available dataset will be resized to the corresponding resolution for each client.

\textbf{B. Medical image classification (different label distributions).} For the breast cancer classification task, we select 50 pieces of data for each client to upload, and the public dataset contains 400 images. For the OCT disease classification task, We selected 1000 pieces of data from both the non training and testing sets.

\textbf{C. Medical time-series classification.} We select 200 pieces of data for each client to upload, and the public dataset contains 600 images.

\section{The Messenger Models}

% \subsection{The Messenger Models in Classification}

For medical image classification, the structure of the messenger model is shown in Table 1. 

% Table generated by Excel2LaTeX from sheet 'Sheet3'
\begin{table}[htbp]
  \centering
  \caption{The structure of the messenger model in the medical image classification. In the messenger body, the stride of each layer is 2. Class represents the category.}
  \resizebox{0.45\linewidth}{!}{
    \begin{tabular}{c|c|c}
    \hline
          & \textbf{Layer} & \textbf{Operation} \\
    \hline
    \multicolumn{1}{c|}{\multirow{5}[2]{*}{Messenger Body}} & Conv2d 3x3-64 & ReLU \\
          & MaxPool 3x3 & - \\
          & Conv2d 5x5-64 & ReLU \\
          & MaxPool 3x3 & - \\
          & Conv2d 7x7-512 & ReLU \\
    \hline
    \multicolumn{1}{c|}{\multirow{2}[2]{*}{Messenger Head}} & Linear-256 & BatchNorm1d+ReLU \\
          & Linear-class & - \\
    \hline
    \end{tabular}%
    }
  \label{tab:addlabel}%
\end{table}%

For medical image segmentation, the structure of the messenger model is shown in Table 2.

% Table generated by Excel2LaTeX from sheet 'Sheet3'
\begin{table}[htbp]
  \centering
  \caption{The structure of the messenger model in the medical image segmentation. In the messenger body, the stride of each conv2d layer is 1 and MaxPool is 2. In the messenger head, the stride of each layer is 2. Class represents the category.}
  \resizebox{0.45\linewidth}{!}{
    \begin{tabular}{c|c|c}
    \toprule
          & \textbf{Layer} & \textbf{Operation} \\
    \midrule
    \multicolumn{1}{c|}{\multirow{8}[2]{*}{Messenger Body}} & Conv2d 3x3-64 & ReLU \\
          & MaxPool 3x3 & - \\
          & Conv2d 5x5-64 & ReLU \\
          & MaxPool 3x3 & - \\
          & Conv2d 7x7-64 & ReLU \\
          & MaxPool 3x3 & - \\
          & Conv2d 7x7-512 & ReLU \\
          & MaxPool 3x3 & - \\
    \midrule
    \multicolumn{1}{c|}{\multirow{4}[2]{*}{Messenger Head}} & ConvTranspose2d 2x2-64 & ReLU \\
          & ConvTranspose2d 2x2-64 & ReLU \\
          & ConvTranspose2d 2x2-64 & ReLU \\
          & ConvTranspose2d 2x2-class & - \\
    \bottomrule
    \end{tabular}%
    }
  \label{tab:addlabel}%
\end{table}%

\section{Future Works} 
Our method has demonstrated its effectiveness in medical classification and segmentation tasks \cite{greenspan2023medical,xie2024trls,xie2022tc,xie2022single, gong2021deformable}. In the future, We will extend our approach to areas such as medical object detection, image super-resolution \cite{xie2023shisrcnet}, 3D reconstruction \cite{luan2023high,luan2021pc,zhai2023towards,luan2024spectrum,wu2024fsc,gong2023progressive}, medical image generation \cite{chen2024federated}, Magnetic Resonance Imaging \cite{chen2014motion}, \textit{etc.}. Moreover, we will also leverage the foundation models \cite{moor2023foundation} into our FL framework to achieve higher performance.

\end{document}